\documentclass[runningheads]{llncs}
\usepackage{graphicx}
\usepackage[T1]{fontenc}

\usepackage{amsmath}
\usepackage{amscd}
\usepackage{amssymb}
\usepackage{color}
\usepackage{graphicx}
\usepackage{wrapfig}
\newcommand{\R}{\mathbb{R}}

\newcommand{\cB}{\mathcal{B}}

\newcommand{\beq}{\begin{equation}}
\newcommand{\eeq}{\end{equation}}

\usepackage{caption}
\usepackage{subcaption}

\begin{document}
\title{\Large\bfseries Geometric Deep Learning: a Temperature Based \\ Analysis 
of Graph Neural Networks}
%
\titlerunning{Geometric Deep Learning: a Temperature Based}
%
\author{Lapenna M.\inst{1}\orcidID{0000-0001-5293-9499} \and
Faglioni F. \inst{2}\orcidID{0000-0002-3327-8840}  \and \\
Zanchetta F.\inst{1}\orcidID{0000-0003-4075-7641} \and
Fioresi R. \inst{1}\orcidID{0000-0003-2294-2755}}

\authorrunning{Faglioni F. et al.}
%
\institute{University of Bologna, Bologna, Italy
\email{rita.fioresi@unibo.it, michela.lapenna4@unibo.it, 
ferdinando.zanchetta2@unibo.it} 
\and
University of Modena, Modena, Italy,
\email{francesco.faglioni@unimore.it}
}
%
%
\maketitle              
\begin{abstract}
We examine a Geometric Deep Learning model as a thermodynamic system treating the weights as 
{non-quantum and non-relativistic} particles. We employ the notion of temperature previously defined in \cite{ffms} and study it in the various layers for GCN
and GAT models. 
Potential future applications of our findings are discussed.
\end{abstract}

\section{\large\bfseries Introduction} \label{intro-sec}

Machine learning and statistical mechanics
share a common root;  starting from the pioneering works by 
Jaynes (\cite{jaynes}) and Hopfield (\cite{hopfield}), up to
the visionary theory of (deep) Boltzmann machines (\cite{boltz}, \cite{dbm}), it is clear there is a common ground and the understanding of statistically inspired machine learning models can bring a new impulse to the field.
The powerful language of statistical mechanics, connecting
the elusive microscopic and measurable macroscopic physical quantities seems the perfect framework to tackle the difficult interpretation questions that the successful deep neural networks present.
Indeed many researchers (see \cite{cs}, \cite{ccs} and refs. therein) have conducted a thermodynamic study, through analogies, of various actors in the most popular algorithms,
Deep Learning above all, and yet such analogies were not able to
fully elucidate the mechanisms of generalization and representability that still elude our understanding.
Along the same vein, new mathematical modeling, inspired by
thermodynamics, brought along new interesting mathematics, see
\cite{barbaresco1}, \cite{marle}, and in particular \cite{deleon}, that seems especially suitable to model the dissipative phenomenon we observe in the SGD experiments.

\medskip
The purpose for our present paper is to initiate this thermodynamic analysis for the Geometric Deep Learning algorithm along the same line of our previous works \cite{ffms} and
\cite{lff} on more traditional Convolutional Neural Networks (CNN). 
We shall treat the parameters of the model as a thermodynamic
system of particles and we exploit the 
sound notion of temperature we have previously given in \cite{ffms} and \cite{lff}.
Then, we study the temperature of the system across layers in
Graph Convolutional Networks (GCN) \cite{kw} 
and Graph Attention Networks (GAT) \cite{v}.

\medskip 
Our paper is organized as follows. In Sec. \ref{thermo-sec}
we briefly recall the correspondence between
thermodynamics concepts and neural networks ones (\cite{ffms}).
In Sec. \ref{exp-sec} we present a Geometric Deep Learning
model on the MNIST Superpixels dataset (\cite{superpixel}) and we study the temperature of layers comparing with 
the behaviour found for the 
CNN architecture in \cite{ffms} and \cite{lff}. In particular, we 
study the dependence of temperature from the two hyperparameters learning rate and batch size at the end of the training, when loss and accuracy have reached their equilibrium values. We also analyse the dynamics of the weights inside a single Graph Convolutional layer.
We consider both a GCN and a GAT models and compare 
the results.
In Sec. \ref{concl-sec} we draw our conclusions and we 
lay foundations for future work.

\medskip

\section{\large\bfseries Thermodynamics
  and Stochastic Gradient Descent}
\label{thermo-sec}

We briefly summarize the thermodynamic analysis and modelling appearing in \cite{cs}, \cite{ffms}, and \cite{lff}. 

\medskip
Stochastic Gradient Descent (SGD) and its variations
(e.g. Adam) are common choices, when performing optimization
in 
Deep Learning algorithms. 

\medskip
Let $\Sigma=\{z_i \,| 1 \leq i \leq N\} \subset \R^D$ denote a
dataset of size $N$, i.e., $|\Sigma|=N$,
$L=(1/N)\sum L_i$ the loss function, with $L_i$
the loss of the $i$-th datum $z_i$ and $\cB$ the minibatch.
The update of the weights $w=(w_k) \in \R^d$ of the chosen model
(e.g., Geometric Deep Learning model), 
with SGD occurs as follows:
\beq\label{time-evol}
w(t+1)=w(t) -\eta \nabla_{\mathcal{B}} L(w), \quad
\hbox{with} \quad 
\nabla_{\mathcal{B}} L := \frac{1}{|\mathcal{B}|}
\sum_{i \in \mathcal{B}}\nabla L_i
\eeq
where $\eta$ denotes the learning rate. 
Equation (\ref{time-evol}) is modelled in \cite{cs} by the 
stochastic ODE (Ito formalism \cite{risken}) 
expressed in its continuous version as:
\begin{equation}\label{time-evol-stoc}
dw(t) = -\eta\nabla L(w)dt +
\sqrt{2\zeta^{-1}D(w)}dW(t)
\end{equation}
where $W(t)$ is the {\sl Brownian motion} term modelling the 
stochasticity of the descent, while $D(w)$ is the {\sl diffusion matrix}, controlling the anisotropy of the diffusivity in the process. The quantity $\zeta=\eta/(2|\mathcal{B}|)$ in \cite{cs}, is called the \textit{temperature}.
It accounts for the ``noise'' due to SGD: 
small minibatch sizes or a high learning rate will increase the
noise in the trajectories of the weights during training.

\medskip
In \cite{ffms}, the time evolution of the parameters is written 
in continuous and discrete version as:
\beq\label{time-evol-cont}
dw(t) = -\eta\nabla_{\mathcal{B}} L(w)dt, \qquad
 w(t+1)=w(t)-\eta\nabla_{\mathcal{B} }L(w)
\eeq
The stochastic behaviour modelled by (\ref{time-evol-cont})
is then accounted for introducing a microscopic definition of temperature mimicking Boltzmann statistical mechanics. We first define the \textit{instantaneous temperature} $\mathcal{T}(t)$ of the system as its kinetic energy  $\mathcal{K}(t)$ divided by the number of degrees of freedom $d$
(in our case the dimension of the weight space) and a constant $k_B > 0$ to obtain the desired units:
\beq\label{inst-temperature}
\mathcal{T}(t) = \frac{\mathcal{K}(t)}{k_B\,d}=
\frac{1}{k_B\,d}\sum_{k=1}^{d} \frac{1}{2}m_k\,v_k(t)^2
\eeq
where $v_k(t)$ is the instantaneous velocity of one parameter, computed as 
the difference between the value of the $k^{\mathrm{th}}$ parameter at one step 
of training and its value at the previous step (the shift in time $\Delta t$ is unitary since we are computing the instantaneous velocity between consecutive steps or epochs):

\beq\label{velo}
v_k(t) = \frac{w_k(t) - w_k(t-1)}{\Delta t}
\eeq

In the formula for the kinetic energy, $m_k$ is the mass of parameter $w_k$ and we set it to $1$. This is because we have do not know 
if the different role of the parameters 
can be modelled through a parallelism with the concept of mass.

The \textit{thermodynamic temperature} is then the time average of 
$\mathcal{T}(t)$:

\beq\label{temperature}
T = \frac{1}{\tau}\int_0^\tau \mathcal{T}(t)\,dt=
\frac{1}{\tau k_B\, d}\int_0^\tau \mathcal{K}(t)=\frac{K}{k_B\, d}
\eeq

where $K$ is the average kinetic energy and $\tau$ is an interval of time long enough to account for small variation in temperature.
In \cite{ffms} we interpret this system as evolving at constant
temperature: at each step the temperature is reset (analogy with system in contact
with heat reservoir). Hence we
do not have a \textit{constant energy} dynamics, as it is commonly referred to in atomic simulations, but we are faced with
a dissipative effect occurring at each step.

The thermodynamic analysis performed in \cite{ffms} and summarized here implies 
that with SGD we have a residual velocity for each particle even after equilibrium is reached. Our system does not evolve according to Newton dynamics and in particular the mechanical energy is not constant.
The fact we maintain a residual temperature at equilibrium
with a constant temperature evolution means that we
achieve a minimum of {\sl free energy}, not of the potential energy i.e. our loss function. This fact is stated in \cite{cs} and is well known among the machine learning and information geometry community (see also \cite{barbaresco1}).

\medskip
Let us summarize the key points of the system dynamics. We have:

-- No costant mechanical energy $K+V$,
where $K=\sum_{k=1}^d (1/2) m_kv_k^2$ is the kinetic energy
and $V=L$ ($L$ the loss function) is the potential energy;

-- No maximization of entropy,

All of this is due to the stochasticity of SGD which is enhanced by small sizes of minibatch and high learning rate,
as we shall elucidate more in our experimental section together
with an analysis of the temperature in the layers.

\section{\large\bfseries Experiments} \label{exp-sec}

In this section we perform
experiments with Geometric Deep Learning models on MNIST Superpixels PyTorch dataset (\cite{superpixel}), in order to test the dependence of the thermodynamic temperature from the hyperparameters learning rate and batch size.
We examine the temperature of different layers and we look at the mean squared velocities of the weights 
of a single layer.
We also analyze some key differences with the findings in \cite{ffms} and in \cite{lff}, where we proposed pruning techniques based on the notion of temperature. 

We choose to investigate the behaviour of two separate 
and important Graph Neural Network (GNN) architectures. First, we implement a model using Graph Convolutional Network (GCN) layers from \cite{kw}. Then, we make a comparison with a Graph Attention Network (GAT) model employing attention mechanism during the convolution \cite{v}. 
We stress that, in the literature, GAT models have outperformed GCN's in classification problems on superpixel images. We consider both models in order to compare the weights' dynamics and reason on the thermodynamic modelling we propose.

\subsection{\bfseries GCN Architecture.} \label{gcn-sec}
\begin{figure}
    \centering
    \includegraphics[width=0.73\textwidth]{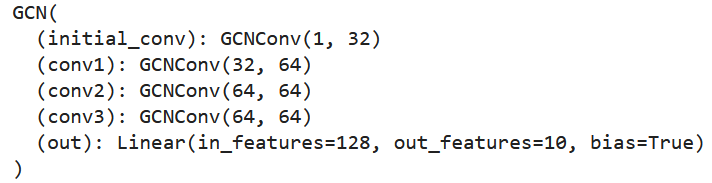}
    \caption{Architecture of our GCN model in PyTorch framework. In the parenthesis next to each layer, the first and second number indicate the embedding dimension of the input and output respectively. The final linear layer takes as input an embedding which is twice the size of the output from the previous layer, due to concatenation of the two pooling operations.} 
    \label{fig:architecture_GCN}
\end{figure}

The architecture we use consists of four GCNConv layers followed by a concatenation of a mean and a max pooling layers and at the end a dense layer
(Fig. \ref{fig:architecture_GCN}). We use $\tanh$ as activation function.

We do not apply batch normalization (\cite{batchnorm}) to 
the convolutional layers, since 
normalizing the weights could bias our experiments (compare with \cite{ffms}).
We optimize the network with SGD with a Cross Entropy loss and Adam optimizer, without any form ofweight regularization. 

We train this model on the MNIST Superpixels dataset obtained in \cite{superpixel}, where the $70.000$ images from the original MNIST dataset were transformed into graphs with $75$ nodes, each node corresponding to a superpixel. We train the model for 600 epochs, starting with a learning rate of $10^{-3}$ and decaying it of a $1/10$ factor every 200 epochs. After training, the model reaches an accuracy of $64 \%$, 
which is worse than the performance obtained on the same dataset in \cite{superpixel}. We think this is due to the fact that our architecture is much simpler than MoNET (\cite{monet}) used in \cite{superpixel}. 
Once equilibrium of accuracy and loss is reached, we further train the model for 100 epochs and we focus our thermodynamic analysis on these last epochs at equilibrium. In particular, to investigate the dependence of temperature from learning rate and batch size, we further train the same equilibrium model by changing either the learning rate or the batch size.

\begin{figure}
     \centering
     \begin{subfigure}{0.4\textwidth}
         \centering
         \includegraphics[width=\textwidth]{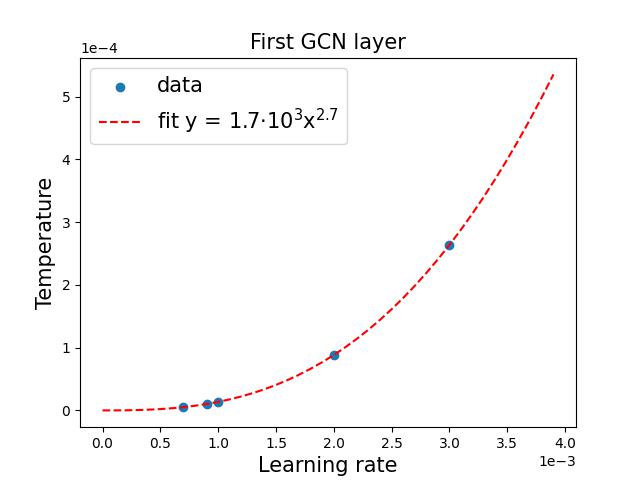}
         \label{fig:First_GCN_lr}
     \end{subfigure}
     \begin{subfigure}{0.4\textwidth}
         \centering
         \includegraphics[width=\textwidth]{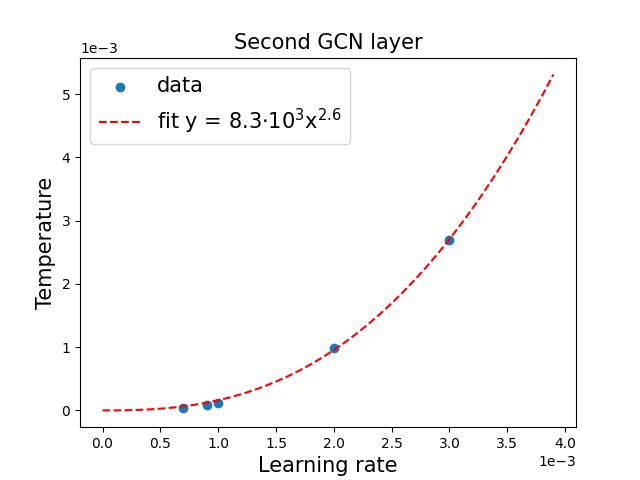}
         \label{fig:Second_GCN_lr}
     \end{subfigure}

     \begin{subfigure}{0.4\textwidth}
         \centering
         \includegraphics[width=\textwidth]{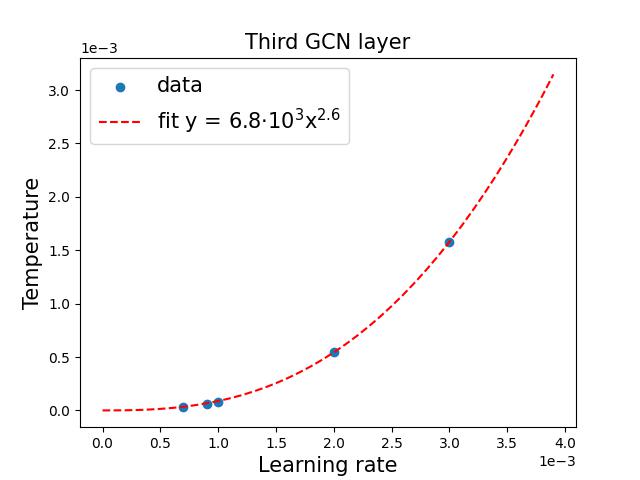}
         \label{fig:Third_GCN_lr}
     \end{subfigure}
     \begin{subfigure}{0.4\textwidth}
         \centering
         \includegraphics[width=\textwidth]{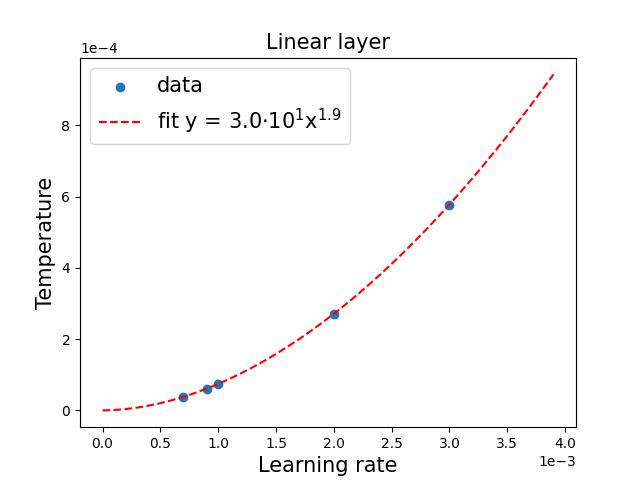}
         \label{fig:Linear_GCN_lr}
     \end{subfigure}
        \caption{Temperature dependence from the learning rate for selected layers of the GCN architecture (the other behaving similarly). The equation resulting from the fit is shown in the top left.}
        \label{fig:temp_eta_GCN}
\end{figure}

\begin{figure}
     \centering
     \begin{subfigure}{0.4\textwidth}
         \centering
         \includegraphics[width=\textwidth]{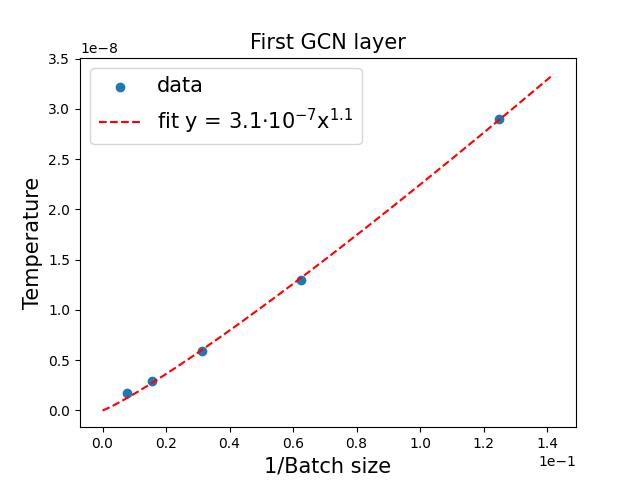}
         \label{fig:First_GCN_b}
     \end{subfigure}
     \begin{subfigure}{0.4\textwidth}
         \centering
         \includegraphics[width=\textwidth]{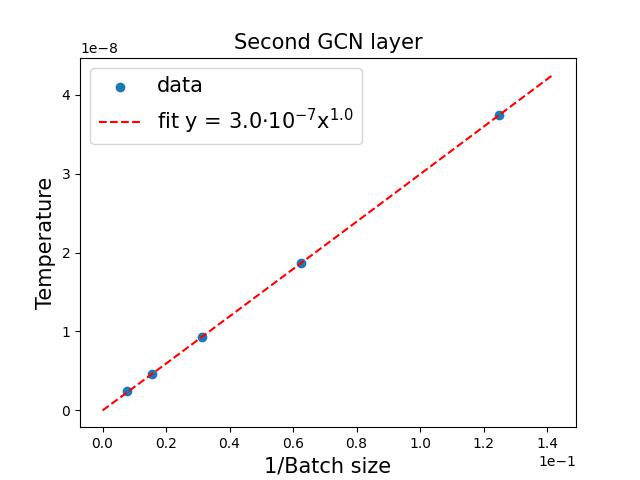}
         \label{fig:Second_GCN_b}
     \end{subfigure}

     \begin{subfigure}{0.4\textwidth}
         \centering
         \includegraphics[width=\textwidth]{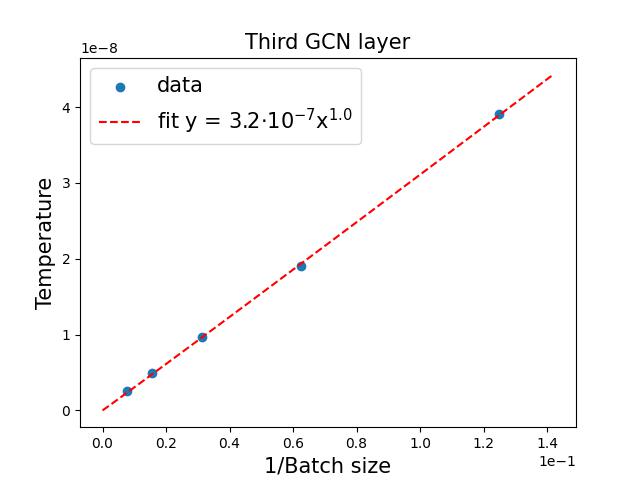}
         \label{fig:Third_GCN_b}
     \end{subfigure}
     \begin{subfigure}{0.4\textwidth}
         \centering
         \includegraphics[width=\textwidth]{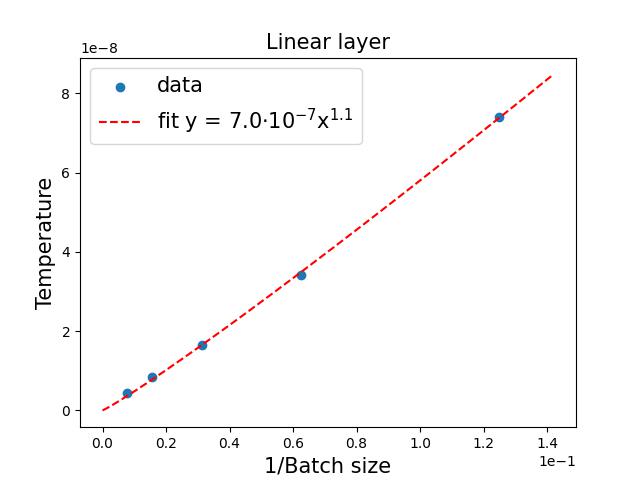}
         \label{fig:Linear_GCN_b}
     \end{subfigure}
        \caption{Temperature dependence from the inverse of the batch size for selected layers of the GCN architecture.}
        \label{fig:temp_beta_GCN}
\end{figure}

In Fig. \ref{fig:temp_eta_GCN} and \ref{fig:temp_beta_GCN}, we show the behaviour of the temperature $T$, as defined
in our previous section, depending on the learning rate $\eta$ and the inverse of the batch size $1/\beta$. We try values of learning rate in the range from $7 \cdot 10^{-4}$ to $3 \cdot 10^{-3}$ (batch size fixed to 32) and values of batch size in the range from 8 to 128 (learning rate fixed). 
We stress that, for each layer of the architecture, the temperature was computed as the mean kinetic energy of the weights 
averaged over the 100 equilibrium epochs.

Despite in the literature (\cite{cs} and refs therein) 
the temperature $\zeta$ (equation \ref{time-evol-stoc}) is commonly believed to behave proportionally to such parameters, we observe quite a different behaviour. As in \cite{ffms}, the dependence of temperature from the learning rate is parabolic for the linear layer, whereas it is almost parabolic for the GCNConv layers, where the exponent of $x$ in the fit is greater than $2$ (see Fig. \ref{fig:temp_eta_GCN}).
As far as the dependence on the batch size, we notice that all layers of the architecture exhibit a linear dependence of temperature from $1/\beta$. This is in line with the results obtained by \cite{cs}, but differs from the ones in \cite{ffms}. Indeed, in \cite{ffms}, the linear dependence of the temperature from $1/\beta$ appears only for the output linear layer, while the other layers exhibit an essential non linearity. However, overall, as for the CNN architecture previously studied (\cite{ffms}, \cite{lff}), we find that, given our thermodynamic definition of temperature, $T$ is not proportional to $\eta/\beta$.

\begin{figure}
     \centering
     \begin{subfigure}{0.48\textwidth}
         \centering
         \includegraphics[width=\textwidth]{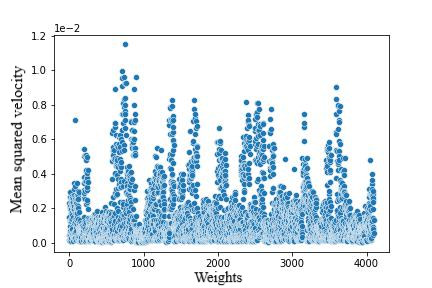}
         \label{fig:2D_Second_GCN}
     \end{subfigure}
     \begin{subfigure}{0.48\textwidth}
         \centering
         \includegraphics[width=\textwidth]{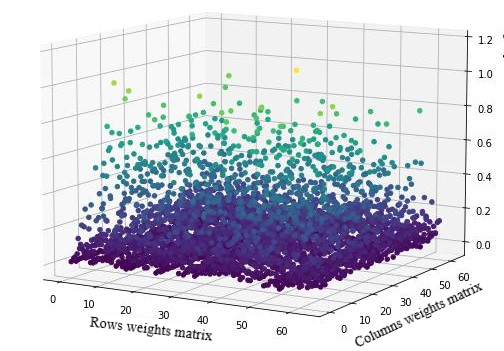}
         \label{fig:3D_Second_GCN}
     \end{subfigure}
         \caption{Mean squared velocity for the weights of the second GCNConv layer. The weight matrix has dimensions $64\times64$ and different rows of the matrix show different temperature. The plot on the left is obtained by flattening the matrix.}
         \label{fig:velocities_GCN}
\end{figure}

Furthermore, if we look at the mean squared velocities of the weights over the epochs, without averaging on the number of weights, we discover quite an interesting behaviour. Inside the same layer, the weights do not show all the same mean squared velocity at equilibrium, but different rows of the weight matrix show completely different thermal agitation (Fig. \ref{fig:velocities_GCN}). This
is similar to what happens in \cite{lff}, where 
we use the concept of temperature to distinguish between ``hot" and ``cold" filters in a CNN layer and we discover that high temperature filters can be removed from the model without affecting the overall performance. We believe
the same reasoning can be applied here for GNN layers and it could imply that some rows of the weight matrix are redundant to the learning. We expect to use this analysis to eliminate useless features 
or to reduce the dimensions of the 
feature embedding and speed up the optimization.

\subsection{\bfseries GAT Architecture.} \label{gat-sec}

\begin{figure}[h!]
    \centering
    \includegraphics[width=0.73\textwidth]{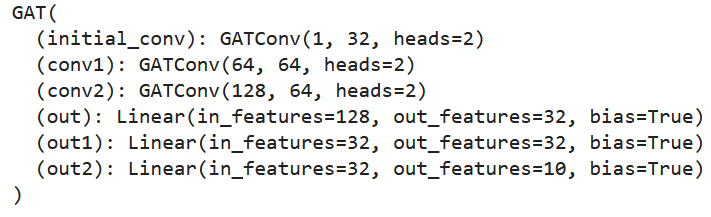}
    \caption{Architecture of our GAT model in PyTorch framework. In the parenthesis next to each layer, the first number and second number indicate the embedding dimension of the input and output respectively. The third parameter indicates the number of heads. Since each GAT in the architecture has $2$ heads, each layer following a GAT has input dimension twice the size of the ouput from the previous GAT layer.} 
    \label{fig:architecture_GAT}
\end{figure}

The architecture we use is inspired by \cite{GATmnist} and consists of three GATConv layers followed by a final mean pooling layer and three dense layers (Fig. \ref{fig:architecture_GAT}). We take ReLU as activation function (\cite{relu}).

\begin{figure}
     \centering
     \begin{subfigure}{0.4\textwidth}
         \centering
         \includegraphics[width=\textwidth]{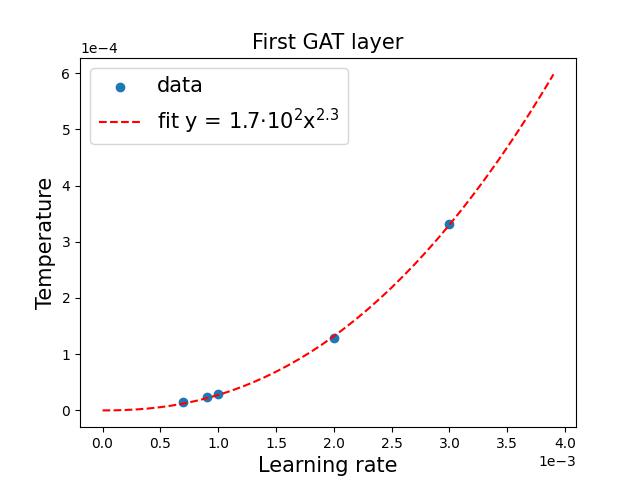}
         \label{fig:First_GAT_lr}
     \end{subfigure}
     \begin{subfigure}{0.4\textwidth}
         \centering
         \includegraphics[width=\textwidth]{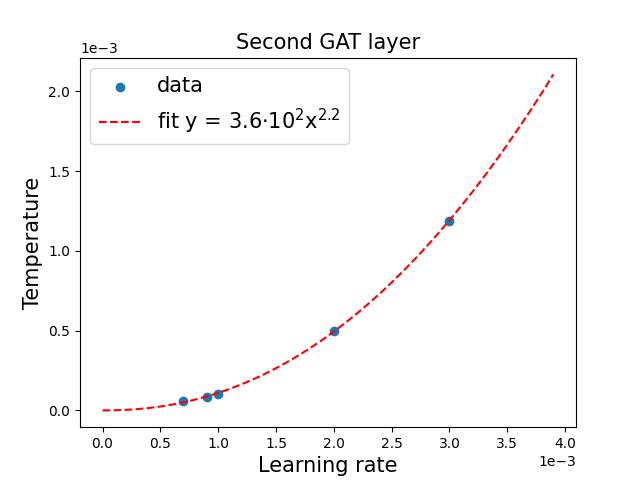}
         \label{fig:Second_GAT_lr}
     \end{subfigure}


     \begin{subfigure}{0.4\textwidth}
         \centering
         \includegraphics[width=\textwidth]{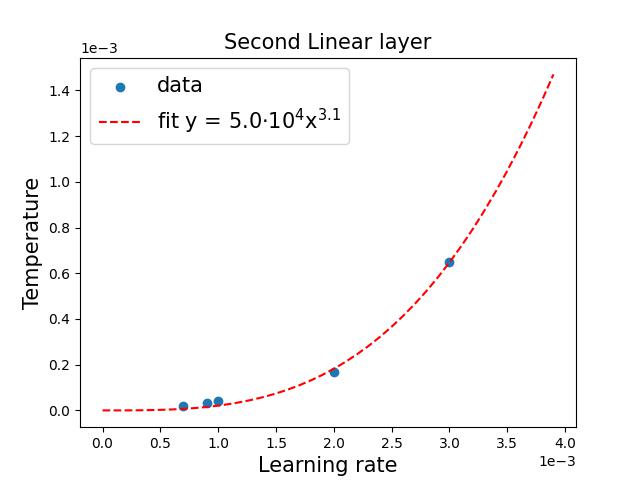}
         \label{fig:Second_Linear_lr}
     \end{subfigure}
     \begin{subfigure}{0.4\textwidth}
         \centering
         \includegraphics[width=\textwidth]{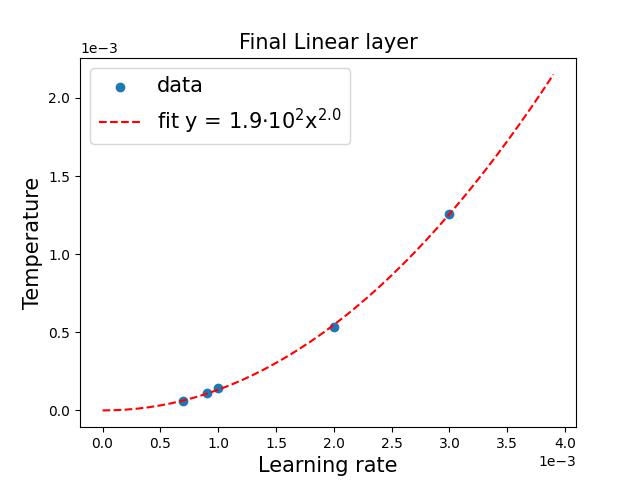}
         \label{fig:Final_Linear_lr}
     \end{subfigure}
        \caption{Dependence of temperature from the learning rate for some layers of the GAT architecture.}
        \label{fig:temp_eta_GAT}
\end{figure}

\begin{figure}
     \centering
     \begin{subfigure}{0.4\textwidth}
         \centering
         \includegraphics[width=\textwidth]{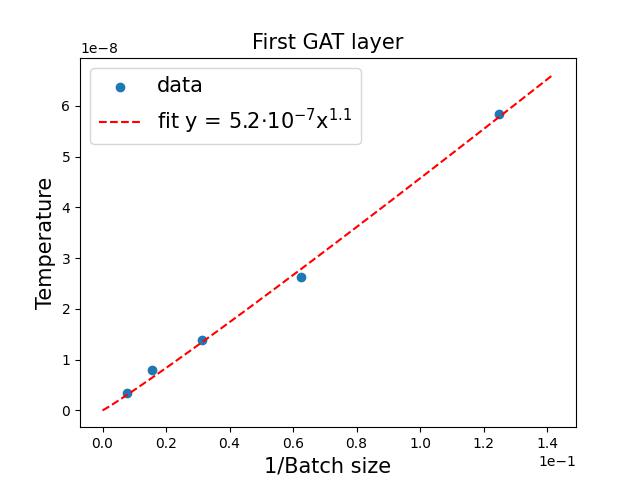}
         \label{fig:First_GAT_b}
     \end{subfigure}
     \begin{subfigure}{0.4\textwidth}
         \centering
         \includegraphics[width=\textwidth]{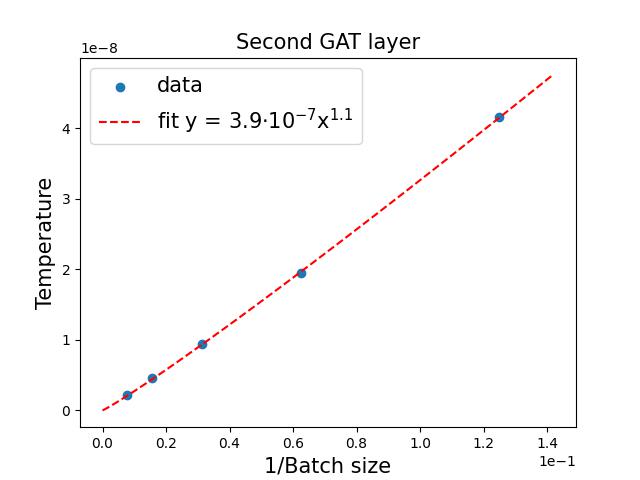}
         \label{fig:Second_GAT_b}
     \end{subfigure}


     \begin{subfigure}{0.4\textwidth}
         \centering
         \includegraphics[width=\textwidth]{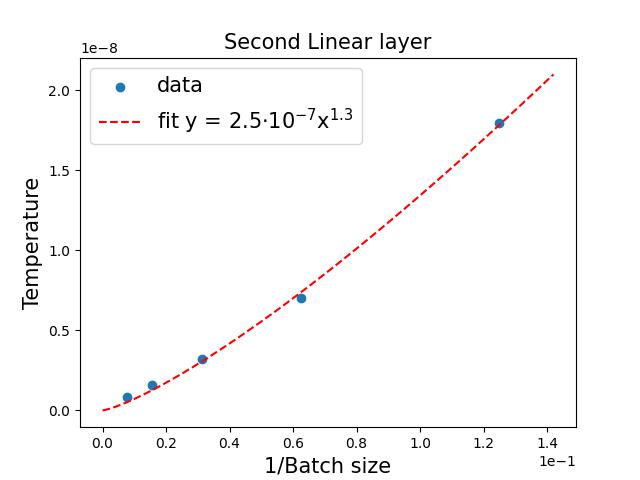}
         \label{fig:Second_Linear_b}
     \end{subfigure}
     \begin{subfigure}{0.4\textwidth}
         \centering
         \includegraphics[width=\textwidth]{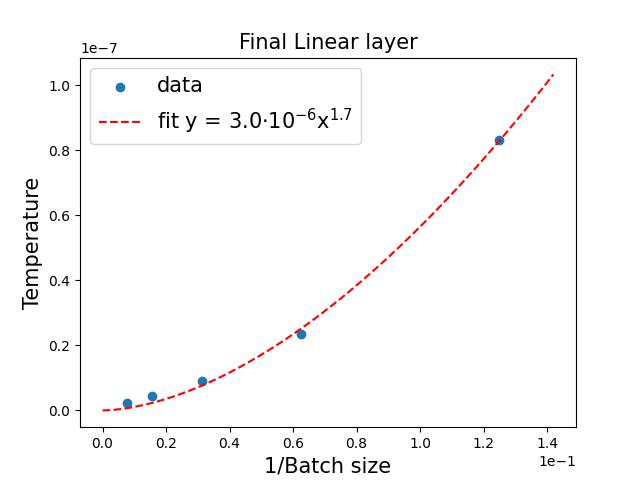}
         \label{fig:Final_Linear_b}
     \end{subfigure}
        \caption{Dependence of temperature from the inverse of the batch size for some layers of the GAT architecture.}
        \label{fig:temp_beta_GAT}
\end{figure}

As for the previous GCN model, we do not apply either batch normalization (\cite{batchnorm}) or dropout (\cite{dropout}) or other forms of regularization to the weights. We optimize the network as described for GCN and we obtain a test set accuracy of $74 \%$. 
To inspect the dependence of temperature on the two hyperparameters $\eta$ and $\beta$, again we train the equilibrium model for other 100 epochs and we restrict the analysis to these final epochs. In Fig. \ref{fig:temp_eta_GAT} and \ref{fig:temp_beta_GAT}, we report the behaviour of the temperature dependence on the learning rate $\eta$ and the inverse of the batch size $1/\beta$ (range of values of the hyperparameters as for the GCN model). Similarly to the model without attention, the dependence of $T$ from $\eta$ is parabolic for the final linear layer and almost parabolic for the other GATConv and linear layers, since for these layers the exponent of $x$ in the fit is greater than $2$ (Fig. \ref{fig:temp_eta_GAT}). Furthermore, the dependence of $T$ from $1/\beta$ is again almost linear for every layer (only the final linear layer shows a more parabolic behaviour).

\section{\large\bfseries Conclusions} \label{concl-sec}

We 
investigate the parallelism between SGD dynamics and thermodynamic systems extending the study in \cite{ffms} and \cite{lff} to Geometric Deep Learning algorithms.
Experiments show that similarly to the Deep Learning setting of CNN architectures (\cite{ffms}, \cite{lff}),
also for Geometric Deep Learning the temperatures of each layer behave independently.
The temperature of the linear layers in the Geometric Deep Learning models considered behaves similarly to that of the linear layers in the CNN models studied in \cite{ffms} and does not exhibit any simple dependence from $\eta$ and $\beta$ as originally assumed in \cite{cs}. 
On the contrary, the temperature of GCN and GAT ``convolutional" layers behaves differently from the one of CNN layers considered in \cite{ffms}: while the dependence of temperature from $\eta$ is again almost parabolic, the one from $\beta$ becomes linear.
Furthermore, we find that different areas of GCNConv and GATConv layers have different temperature (see Fig. \ref{fig:velocities_GCN}) as observed for the filters of a CNN. 
This suggests a future technique of parameter pruning, based on temperature, as in \cite{lff}, that may help speed up the optimization and make it more effective.
More mathematical and physical modelling is needed to further
advance in this direction.

%
%
\bibliographystyle{splncs04}

\begin{thebibliography}{8}

\bibitem{boltz} D. H Ackley, G. E Hinton, and T. J Sejnowski. 
\textit{A learning algorithm for Boltzmann machines.} Cognitive science, 9(1):147-169, 1985.

\bibitem{amari} Shun-Ichi Amari. 
\textit{Natural gradient works efficiently in learning.} 
Neural computation, 10(2):251-276, 1998.

\bibitem{barbaresco1} Fr\'ed\'eric Barbaresco. 
\textit{Lie group statistics and Lie group 
machine learning based on Souriau Lie groups thermodynamics 
and Koszul-Souriau-Fisher metric: new entropy
definition as generalized Casimir invariant function in coadjoint representation.}
Entropy, 22(6):642, 2020.


\bibitem{cs} Pratik Chaudhari and Stefano Soatto. 
\textit{Stochastic gradient descent performs variational inference, converges to limit cycles for deep networks.} 
2018 Information Theory and Applications Workshop (ITA), pages 1-10. 

\bibitem{ccs}
P. Chaudhari, A. Choromanska, S. Soatto, Y. LeCun et al.
\textit{Entropy-sgd: Biasing gradient descent into
wide valleys.}
International Conference on Learning Representations ICLR 2017, 1611.01838.
  
\bibitem{cfs} R. Fioresi, P. Chaudhari and S. Soatto. 
\textit{A geometric interpretation of stochastic gradient descent using diffusion metrics.} Entropy, 22(1):101, 2020.

\bibitem{ffms}
R. Fioresi, F. Faglioni, F. Morri and L. Squadrani.
\textit{On the thermodynamic interpretation of deep learning systems.}
Geometric Science of Information: 5th International
Conference, 2021.
  
\bibitem{hopfield} J. Hopfield J. 
\textit{Neurons with graded response have collective computational properties like those of two-state neurons.} 
PNAS, 81(10):3088-3092.

\bibitem{jaynes}
Edwin T Jaynes. \textit{Information theory and statistical mechanics.} Physical review, 106(4):620, 1957.

\bibitem{kri2012}
A. Krizhevsky, I. Sutskever and G. E Hinton. 
\textit{Imagenet classification with deep convolutional neural networks.} 
Advances in neural information processing systems, 25:1097-1105, 2012.

\bibitem{lff}
M. Lapenna, F. Faglioni and R. Fioresi
\textit{Thermodynamics Modeling of Deep Learning
Systems}, preprint, Frontiers in Physics, 2023.

\bibitem{lecun2015}
Yann LeCun, Yoshua Bengio, and Geoffrey Hinton. \textit{Deep learning}.
Nature, 521(7553):436-444, 2015.

\bibitem{marle} Charles-Michel Marle. 
\textit{From tools in symplectic and Poisson geometry to J.
M. Souriau's theories of statistical mechanics and thermodynamics.} Entropy, 18(10):370, 2016.

\bibitem{dbm}
Ruslan Salakhutdinov and Geoffrey Hinton. 
\textit{Deep Boltzmann machines.}
In Artificial intelligence and statistics, pages 440-455. PMLR, 2009.

\bibitem{deleon}
A. Anahory Simoes, M. De Leon, M. Lainz Valcazar and D. M. De Diego. 
\textit{Contact geometry for simple thermodynamical systems
with friction.} 
Proceedings of the Royal Society A, 476(2241):20200244, 2020.

\bibitem{wt}
Max Welling and Yee Whye Teh.
\textit{Bayesian learning via stochastic gradient langevin dynamics.} In International Conference on Machine Learning, 2011.

\bibitem{batchnorm}
Sergey Ioffe and Christian Szegedy.
\textit{Batch Normalization: Accelerating Deep Network Training by Reducing Internal Covariate Shift.} Proceedings of the 32nd International Conference on Machine Learning, PMLR 37:448-456, 2015.

\bibitem{superpixel}
F. Monti, D. Boscaini, J. Masci,
E. Rodol{\`{a}} et al. 
\textit{Geometric deep learning on graphs and manifolds using mixture model CNNs}
2017 IEEE Conference on Computer Vision and Pattern Recognition (CVPR), 10.1109/CVPR.2018.00335.

\bibitem{monet}
M. Gou, F. Xiong, O. Camps and M. Sznaier.
\textit{MoNet: Moments Embedding Network}
2018 IEEE/CVF Conference on Computer Vision and Pattern Recognition (CVPR), 1611.08402.

\bibitem{GATmnist}
P. H. C. Avelar, A. R. Tavares, T. L. T. da Silveira et al. 
\textit{Superpixel Image Classification with Graph Attention Networks.} 
33rd SIBGRAPI Conference on Graphics, Patterns and Images (SIBGRAPI), 2020, pp. 203-209, 10.1109/SIBGRAPI51738.2020.00035.

\bibitem{relu}
Vinod Nair and Geoffrey E. Hinton. 
\textit{Rectified Linear Units Improve Restricted Boltzmann Machines.} Proceedings of the 27th International Conference on Machine Learning (ICML-10), June 21-24, 2010.

\bibitem{dropout}
N. Srivastava, G. Hinton, A. Krizhevsky et al. 
\textit{Dropout: A Simple Way to Prevent Neural Networks from Overfitting.} The Journal of Machine Learning Research, Volume 15, Issue 1, pp 1929–1958.

\bibitem{kw}
Thomas N. Kipf and Max Welling.
\textit{Semi-Supervised Classification with Graph Convolutional Networks}
International Conference on Learning Representations ICLR 2017, url = https://openreview.net/forum?id=SJU4ayYgl.

\bibitem{risken}
Hannas Risken.
\textit{ The Fokker-Planck Equation,
Methods of Solution and Applications}, Springer, 1996.

\bibitem{v}
Veličković, 
Bengio, Yoshua et al.
\textit{Graph Attention Networks}
International Conference on Learning Representations ICLR 2018, 10.40550/ARXIV.1710.10903.

\end{thebibliography}
%

\end{document}